\documentclass[a4paper,twoside]{article}

\usepackage{epsfig}
\usepackage{subcaption}
\usepackage{calc}
\usepackage{amssymb}
\usepackage{amstext}
\usepackage{amsmath}
\usepackage{amsthm}
\usepackage{multicol}
\usepackage{pslatex}
\usepackage{apalike}
\usepackage{algorithm2e}
\usepackage[bottom]{footmisc}

\usepackage{dsfont}
\usepackage{comment}
\usepackage{booktabs}
\usepackage{algorithm2e}
\RestyleAlgo{ruled}
\usepackage{amsmath}
\usepackage{amsfonts}
\DeclareMathOperator*{\argmax}{arg\,max}
\DeclareMathOperator*{\argmin}{arg\,min}
\usepackage{subcaption}
\usepackage{hyperref}
\usepackage{color}

\usepackage[labelfont=bf]{caption}
\urldef{\myurl}\url{https://archive.ics.uci.edu/datasets}

\usepackage{SCITEPRESS}     

\begin{document}

\title{Enhanced Data-Driven Product Development via Gradient Based Optimization and Conformalized Monte Carlo Dropout Uncertainty Estimation\thanks{Accepted at the 18th International Conference on Agents and Artificial Intelligence (ICAART 2026). Cite as: \emph{Thomas Nava, A., Johny, L., Azzalini, F., Schneider, J., and Casanova, A. (2026). Enhanced data-driven product development via gradient-based optimization and conformalized Monte Carlo dropout uncertainty estimation. In Proceedings of the 18th International Conference on Agents and Artificial Intelligence (ICAART)}}}


\author{\authorname{
Andrea Thomas Nava\sup{1}, 
Lijo Johny\sup{2}, 
Fabio Azzalini\sup{3}, 
Johannes Schneider\sup{2}, 
Arianna Casanova\sup{2}}
\affiliation{\sup{1}ETH Zurich, Zurich, Switzerland}
\affiliation{\sup{2}University of Liechtenstein, Vaduz, Liechtenstein}
\affiliation{\sup{3}Politecnico di Milano, Milan, Italy}
\email{\{lijo.johny, johannes.schneider, arianna.casanova\}@uni.li, 
andrea.nava@ethz.ch, fabio.azzalini@polimi.it}
}


\abstract{Data-Driven Product Development (DDPD) leverages data to learn the relationship between product design specifications and resulting properties. To discover improved designs, we train a neural network on past experiments and apply Projected Gradient Descent to identify optimal input features that maximize performance. 
Since many products require simultaneous optimization of multiple correlated properties, our framework employs joint neural networks to capture interdependencies among targets. Furthermore, we integrate uncertainty estimation via \emph{Conformalised Monte Carlo Dropout} (ConfMC), a novel method combining Nested Conformal Prediction with Monte Carlo dropout to provide model-agnostic, finite-sample coverage guarantees under data exchangeability. Extensive experiments on five real-world datasets show that our method matches state-of-the-art performance while offering adaptive, non-uniform prediction intervals and eliminating the need for retraining when adjusting coverage levels.}

\onecolumn
\maketitle
\normalsize
\setcounter{footnote}{0}
\vfill

\section{Introduction}
Innovation in design has long driven progress across diverse fields, from engineering to architecture. Traditionally, design relied on human intuition, expert knowledge, and iterative prototyping. With the rapid growth of data and advances in machine learning, however, data-driven product development (DDPD) has emerged as a powerful paradigm for guiding design through data-informed insights and optimization \cite{feng2020data,killoran2017generating}. We formalize the DDPD problem by assuming an unknown function $f$ that maps product features $x \in \mathbb{R}^d$ to a property of interest $f(x) \in \mathbb{R}$. The goal is to identify the maximizer:
\begin{align}
    x^* = \argmax_{x} f(x).
    \label{eq:optim}
\end{align} 
In practice, only noisy evaluations of $f(x)$ are available. Our approach first models 
$f$ using a multi-layer perceptron (MLP) and then exploits its differentiability with respect to input features to perform gradient based optimization for proposing improved designs. Many applications involve multiple correlated properties \cite{babutzka2019machine}. To handle this, we employ multi-output MLPs, which jointly model target properties and capture correlations during gradient based search. To further enhance user control and limit the extent of extrapolation,
we also introduce the option to constrain the search space within a feasible range
using Projected Gradient Descent \cite{pgd}.

Uncertainty quantification (UQ) is critical in industrial settings to inform decision-making. High uncertainty in predictions can indicate that a suggested prototype may be unreliable. In this paper, we also provide UQ through prediction intervals and coverage estimates using our novel \emph{ConfMC} method, which combines Monte Carlo Dropout \cite{gal2016dropout} with Nested Conformal Prediction \cite{Gupta_2022} to deliver model-agnostic, finite-sample guarantees.

Our contributions can be summarized as follows: (i) \emph{Neural and Gradient based DDPD}: This work operationalizes the use of Projected Gradient Descent on Multi-Output Artificial Neural Networks (ANNs) to address issues of extrapolation and target correlation in DDPD tasks; (ii) \emph{Uncertainty Quantification}: ConfMC provides valid, calibrated UQ for point predictions; (iii) \emph{Flexibility and user support}: The framework enables users to guide the discovery process and access intuitive, easy-to-understand uncertainty quantification tools.

\section{Related Work}
Data-driven product development (DDPD) leverages historical experimental or simulation data to guide design through data-informed optimization, enabling efficient exploration of complex design spaces \cite{feng2020data,shao2020}.

A common framework in DDPD is Bayesian Optimization (BO) \cite{frazier2018tutorial}. It constructs a surrogate model of the objective function using Gaussian process regression to quantify uncertainty and employs an acquisition function to guide the sampling process. However, BO can be impractical in industrial settings, as it typically requires repeated evaluations, which can be too expensive to obtain in many applications \cite{binois2022survey}.

Population-based algorithms, such as evolutionary and particle swarm methods \cite{puatruaușanu2024,meess2024}, offer flexibility in exploring complex design spaces but remain computationally expensive due to the large number of evaluations required.

Recent work has increasingly focused on gradient based optimization. For example, \cite{royster2023} provide a methodological analysis of achieving targeted performance through design variable adjustments, while \cite{gufler2021} review approaches for flexible multibody systems. Other studies have explored specific industrial applications, including \cite{thelen2022,krsikapa2026}. Following this line, our approach leverages neural networks to approximate the objective function and employs projected gradient descent for iterative optimization in the input space.

Deep generative models, such as GANs and VAEs \cite{killoran2017generating,encoding_vaes,truss,yonekura2024,ghasemi2024}, have also been explored to capture complex design correlations and generate novel candidates. However, their reliance on large datasets limits their applicability in contexts where only limited historical or experimental data are available.

Accurately representing uncertainty is increasingly recognized as essential for reliable optimization and decision-making, particularly when data are scarce. Precise uncertainty might even help to improve accuracy \cite{sch25} or help in constructing causal models\cite{de25ca}. Traditional methods, including Bayesian Neural Networks \cite{mackay1992}, Monte Carlo (MC) Dropout \cite{gal2016dropout}, and Deep Ensembles \cite{lakshminarayanan2017}, often lack finite-sample guarantees and can be computationally demanding. On the other hand, conformal Prediction (CP) \cite{angelopoulos2022gentle} addresses these limitations by providing distribution-free, finite-sample coverage guarantees. Building on these advances, our Conformalised Monte Carlo Dropout (ConfMC) integrates MC Dropout with Nested CP to produce flexible, non-uniform prediction intervals and enable coverage adjustment without retraining.

\section{Background}
\subsection{Gradient Descent and Projected Gradient Descent}
Gradient Descent \cite{ruder2017overview} is a widely used optimization algorithm, commonly employed to train Deep Neural Networks (DNNs) by iteratively updating the network parameters $\theta$ to minimize the expected loss over a data-generating distribution $\mathcal{P}$:
\begin{align}
\theta^* = \argmin_{\theta} \ \mathbb{E}_{(X,Y) \sim \mathcal{P}}[\textit{L}(Y, \hat{f}_{\theta}(X))],
\end{align}
where $\textit{L}$ denotes the loss function, $Y$ is the true response, and $\hat{f}_{\theta}(X)$ is the predicted response.

In practice, the expectation is replaced by an empirical average,  and therefore the $i$-th iteration of gradient descent can be written as:
\begin{align}
\theta_{i+1} \leftarrow \theta_{i} - \eta \nabla_{\theta}\Big(\frac{1}{N} \sum_{i=1}^N L(y_i, \hat{f}_{\theta}(x_i)) \Big)
\end{align}
where $\eta$ is the learning rate and $\nabla_{\theta}\Big(\frac{1}{N} \sum_{i=1}^N L(y_i, \hat{f}_{\theta}(x_i)) \Big)$ is the gradient of the empirical loss function with respect to the DNN's parameters.

In many optimization settings, the parameter space is subject to constraints that restrict the solution to a feasible set $\Theta$. In such cases, the optimization problem can be formulated as:
\begin{align}
\theta^* = \argmin_{\theta \in \Theta} \ \mathbb{E}_{(X,Y) \sim \mathcal{P}}[\textit{L}(Y, \hat{f}_{\theta}(X))].
\end{align}
To handle these constraints, Projected Gradient Descent (PGD) \cite{pgd} extends standard gradient descent by incorporating a projection step that enforces feasibility at each iteration. After computing a gradient update, the intermediate solution is projected back onto the feasible set $\Theta$ using the projection operator $P_\Theta$, which returns the closest point within $\Theta$:
\begin{align}
\theta_{i+1} \leftarrow P_{\Theta}( \theta_{i+1}) := \argmin_{\theta \in \Theta} ||\theta - \theta_{i+1}||_2.
\end{align}

\subsection{Monte Carlo Dropout}

Dropout \cite{JMLR:v15:srivastava14a} is a regularization technique applied during the training of a neural network. During training, individual neurons are randomly set to zero with a probability $p$, which prevents the network from relying too heavily on specific neurons.

Monte Carlo (MC) Dropout is an uncertainty quantification method for DNNs that can be interpreted as a Bayesian approximation \cite{gal2016dropout}. It operates by applying dropout at inference time and performing multiple stochastic forward passes for the same input. The resulting collection of predictions forms a heuristic predictive distribution for the corresponding output, providing an estimate of the model’s uncertainty.

\subsection{Nested Conformal Prediction}
Conformal Prediction (CP) \cite{angelopoulos2022gentle} is a distribution free, model-agnostic procedure to generate marginally valid prediction intervals with finite sample theoretical guarantees. This framework has been used to estimate the predictive uncertainty in DNNs particularly in the context of low-dimensional medical data \cite{karimi2023quantifying}. A \textit{marginally} valid prediction interval must satisfy that:
\begin{align}
\mathbb{P} \Big( Y \in C(X_\text{new}) \Big) \geq 1-\alpha,
\label{eq:marginal}
\end{align}
where $C(X_{\text{new}})$ is the prediction interval for a new test input $X_{\text{new}}$ and $\alpha$ is the chosen error rate. The only assumption required by CP is \textit{exchangeability}\footnote{A collection of random variables is exchangeable if their joint distribution is invariant under permutations. See \cite{kuchibhotla2021exchangeability} for details.}. The construction of a CP interval relies on a \textit{non-conformity score}, which measures how much a given observation $(X,Y)$ deviates from examples observed during training. A common choice of score is the absolute residual
\begin{align}
s(X,Y) = |Y - \hat{f}(X)|,
\end{align}
where $\hat{f}$ is the trained prediction model. The prediction interval includes all values of $Y$ that are sufficiently conforming to prior observations.

A widely used variant, \emph{Split (Inductive) Conformal Prediction} \cite{10.1007/3-540-36755-1_29}, uses a calibration set, separate from the model-fitting process, to adjust the intervals to achieve the desired coverage. In this setting, a $(1-\alpha)\cdot 100\%$ prediction interval for a new test input $X_{\text{new}}$ contains all $Y$ such that the non-conformity score $s(X_{\text{new}}, Y)$ does not exceed the $(1-\alpha)\cdot 100\%$ quantile of the empirical distribution of non-conformity scores (as computed from the
calibration set):
\begin{align}
C(X_{\text{new}}) = \{Y: s(X_{\text{new}}, Y) < Q(S, 1-\alpha)\},
\end{align}
where $Q(S, 1-\alpha)$ denotes the corresponding quantile of the empirical distribution of non-conformity scores $S$.

A generalization of CP is provided by the \emph{Nested Sets interpretation} \cite{Gupta_2022}. Here, a nested sequence of prediction intervals $\{C_t(X)\}_{t \in \mathcal{T}}$ is constructed such that each interval contains the previous one, i.e., for every $t_1 \le t_2$, $C_{t_1}(X) \subseteq C_{t_2}(X)$. Using the calibration set, one can select the \textit{smallest} interval in the sequence that achieves the desired coverage $(1-\alpha)$. Again, assuming exchangeability between data points allows
one to expect that also on new test points we will achieve the same coverage rate (up
to random fluctuations). 



\section{Proposed Methodology}
A trained neural network can be viewed as a parameterized function $\hat{f}$ representing the property of interest given prototype specifications. Importantly, since the network is differentiable with respect to its inputs, we can optimize prototype features directly using gradient based methods. Unlike standard neural network training, where gradients are computed with respect to model parameters, here we compute gradients with respect to the input features themselves. This allows us to efficiently navigate the design space. 

For instance, to identify a prototype $x \in \mathbb{R}^d$
 that maximizes a target property, we solve the following optimization problem after training the network on that property:
\begin{equation}
 x^* = \argmax_{x} \hat{f}(x).
\end{equation}
We allow for more flexibility by allowing the user to specify a desired target $t$ (we can recover the above solution by setting $t$ very large). The optimization problem becomes:

\begin{equation}
 x^* = \argmin_{x} ||\hat{f}(x) - t ||_2.
\end{equation}
We try to solve the above problem using gradient descent. Let us denote by $x_i$ the current prototype, by $\nabla_x G(x_i)$ the gradient of $G = ||\hat{f}(x) - t ||_2$ with respect to $x$ and evaluated at $x_i$ and by $\eta$ the learning rate. The gradient steps will then look like:
\begin{equation}
x_{i+1} \leftarrow x_i - \eta \nabla_x G(x_i).
\end{equation}
We incorporate two additional constraints into the procedure. First, to prevent extreme extrapolation, the search is restricted to user-specified ranges for the input features. Second, only selected input features are updated, while fixed features, such as those representing testing conditions, remain unchanged. To enforce the first constraint, we implement Projected Gradient Descent with a projection operator $P$, which, after each gradient step, ensures the prototype lies within the allowed ranges, projecting it to the nearest boundary if necessary. Formally, the user-defined ranges specify a $d$-dimensional hyper-rectangle: $[l_1, u_1] \times [l_2, u_2] \times ... \times [l_d, u_d]$, 
where $l_i$ and $u_i$ represent the lower and upper bound for the value of feature $i$, respectively.
An orthogonal projection onto a hyper-rectangle simply sets any coordinate outside its user-specified range $[l_i, u_i]$ to the nearest boundary $l_i$ or $u_i$, depending on whether it falls below or above the allowed range.  To enforce the second constraint, i.e., updating only selected features, we define a gradient mask. Let $F_U$ denote the set of features to update and $F_U^C$
 the features to keep fixed. The mask $m$ is defined as:
\begin{equation}
\forall \ i = 1, ..., d: \ \text{m}(i)=
    \begin{cases}
        1 & \text{if } x_i \in F_U \\
        0 & \text{if } x_i \in F_U^C.
    \end{cases}
\end{equation}
Multiplying the gradient vector element-wise by $m$ ensures that updates occur only along the desired directions in $F_U$. A schematic of the gradient based search procedure is provided in Algorithm \ref{algo:pgd}. To reduce the risk of convergence to local optima, the search is repeated from multiple random starting prototypes, with only the solution yielding the lowest loss being retained.

\begin{algorithm}[h!]
    \caption{Projected Gradient Descent search for DDPD}
  \label{algo:pgd}
  \SetKwInput{Input}{Input}
  \SetKwInOut{Output}{Output}

  \Input{%
    $n$: Number of iterations, $\eta$: Learning rate, $m$: Mask, $f$: Neural Network, $x$: starting prototype, $H$: Constraints.
  }
  \Output{%
    $x$: Proposed prototype.
  }
  \BlankLine
  \For{$i \leftarrow 1$ \KwTo $n$}{%
    \tcp{Compute NN Prediction for current prototype}
    output $\leftarrow f(x)$\;
    \tcp{Compute Loss with respect to target}
    $L(x) \leftarrow ||f(x) - t||_2$\;
    \tcp{Compute masked Gradient}
    gradient $\leftarrow (\nabla_x L(x)) \cdot m$\;
    \tcp{Normalize Gradient}
    gradient $\leftarrow$ gradient$/||$gradient$||_2$\;
    \tcp{Gradient Step}
    $x \leftarrow x - \eta \cdot$ gradient\;
    \tcp{Project back if outside of constraints}
    \If{$x \not\in H$}{%
        $x \leftarrow P(x)$\;
    }
  }
  \BlankLine
  \tcp{Return the Proposed Prototype}
  \KwRet $x$\;
\end{algorithm}

\subsection{Extension to Multi-Target Search}

In practice, a product often needs to satisfy multiple target properties simultaneously. A natural extension of our approach is to jointly model multiple targets, allowing the search to optimize all properties at once. Using neural networks, this is straightforwardly implemented with a multi-output architecture. For illustration, consider optimizing a prototype for two properties. After training, the network can be represented as a vector-valued function $\hat{f}$. The corresponding objective becomes:
\begin{equation}
x^* = \argmin_{x}
 ||\hat{f}(x)_1 - t_1 ||_2 + \omega \cdot ||\hat{f}(x)_2 - t_2 ||_2,
\end{equation}
where $\hat{f}(x)_i$ denotes the i-th model output, $t_i$ the corresponding target and $\omega$ user-defined weight controlling the relative importance of the second property. 

\section{Uncertainty Quantification (UC)}
\subsection{Conformalized Monte Carlo Dropout (ConfMC)}

We propose to combine the nested sets framework of Conformal Prediction with Monte Carlo dropout to obtain marginally valid prediction intervals. The need for the conformalization step comes from the fact that the standard Monte Carlo dropout predictive distribution has no guarantee to be correct and hence a potential prediction interval constructed by taking quantiles of this distribution is probably not valid. In other words, it is likely that the prediction interval formed by:
\begin{equation}
    C(X) = [Q(\alpha/2), Q(1-\alpha/2)],
\end{equation} 
where $Q(t)$ denotes the $t$-th quantile of the MC predictive distribution $Y|X$, does not achieve the correct coverage rate. Therefore we start from the naive MC predictive distribution but instead of choosing simply the nominal $\alpha$ level, we construct prediction intervals for a sequence of values $t$ of the form $[Q(t/2), Q(1-t/2)]$, compute the coverage on a held-out calibration set and end up picking the value $t$ that leads to the desired coverage (denoted as $\hat{t}$). Finally, the prediction interval for a new point $X_\text{new}$ will be formed by:
\begin{equation}
    [Q(\hat{t}/2), Q(1-\hat{t}/2)].
\end{equation} 

Finally, in a similar spirit as in Conformal Quantile Regression \cite{romano2019conformalized}, we notice that since our method starts from quantile rather than mean predictions, it is adaptive and does not provide constant-width prediction intervals like standard CP based on absolute residuals. 

\section{Empirical Validation of Uncertainty Quantification}
\subsection{Experimental Set-up}
We conducted a comprehensive experimental evaluation to compare the performance of our proposed uncertainty quantification method (\textbf{ConfMC}) against several existing methods. The comparison was performed on five publicly available datasets commonly used for UQ research (see Table \ref{tab:datasets}).
\begin{table*}[ht!]
\caption{Datasets used for experimental validation.}
    \label{tab:datasets}
    \vspace{8pt}
    \centering
    \begin{tabular}{lcc}
        \hline
        \textbf{Dataset} & \textbf{Features} & \textbf{Rows} \\
        \hline
    Concrete \cite{concrete} & 8 & 1030 \\
        Kernel \cite{kernel} & 14  & 241600\\
        Popularity \cite{popularity} & 59 & 39644 \\
        Superconduct. \cite{superconductivty} & 81 & 21263 \\
        BlogData \cite{blog} & 280 & 52397
    \end{tabular}
\end{table*}

For each dataset, we employed the following UQ methods for comparison: (i) \textbf{Standard Conformal Prediction (CP):} Conformal Prediction with absolute residuals as non-conformity score; (ii) \textbf{Conformal Quantile Regression (CQR) \cite{romano2019conformalized}:} Conformal Prediction method based on quantile regression; (iii) \textbf{NGBoost (NGB) \cite{duan2020ngboost}:} Natural Gradient Boosting which provides a probabilistic prediction by making parametric assumption on the data distribution; (iv) \textbf{Monte Carlo (MC) Dropout:} Standard Monte Carlo dropout; (v) \textbf{Conformalised Monte Carlo Dropout (ConfMC) (ours):} Nested Conformal Prediction around Monte Carlo dropout.

We empirically validate our uncertainty quantification methods and check that the prediction intervals achieve a nominal specified coverage of $(1-\alpha) = 0.8$. As recommended by \cite{Gupta_2022},  we repeat the experiment $B=20$ times for each of the datasets to assess the variability of the empirical coverage and width of the prediction intervals. For each trial, we independently draw $N=1000$ random data points (uniformly without replacement) from the full dataset. The data points are split as follows: $n_{train}=562$ for training, $n_{cal}=188$ for calibration and $n_{test}=250$ for testing.
We check the coverage by computing the \emph{Average Empirical Coverage} (AEC) on the test set:
\begin{align}
AEC = \frac{1}{B} \sum_{b=1}^B \left( \frac{1}{n_{test}} \sum_{i=1}^{n_{test}} \mathds{1}\{Y_i^b \in \hat{C^b}(X_i^b)\} \right),
\end{align}
and the \emph{Average Interval Width} (AIW) as:
\begin{align}
AIW = \frac{1}{B} \sum_{b=1}^B \left( \frac{1}{n_{test}} \sum_{i=1}^{n_{test}} \text{width}(\hat{C^b}(X_i^b)) \right).
\end{align}

Furthermore, we propose to visually check the adaptivity of the UQ methods since mean coverage and mean width do not give us information about conditional coverage. We believe that the latter is especially desired in DDPD situations where we wish to have different uncertainty estimates for different prototypes. 

\subsection{Results}
We report the results of the empirical validation in Tables \ref{tab:res_1} and \ref{tab:res_2}. As shown, the MC approach does not achieve the desired coverage. Similarly, the empirical coverage of NGB exhibits large deviations from the target coverage of $80\%$, indicating that it fails to produce well-calibrated prediction intervals. In contrast, the conformal approaches, including our proposed ConfMC method,perform consistently well, achieving the target $80\%$ coverage across most datasets. 

\begin{figure*}[h!]
\centering
\includegraphics[width=0.8\linewidth]{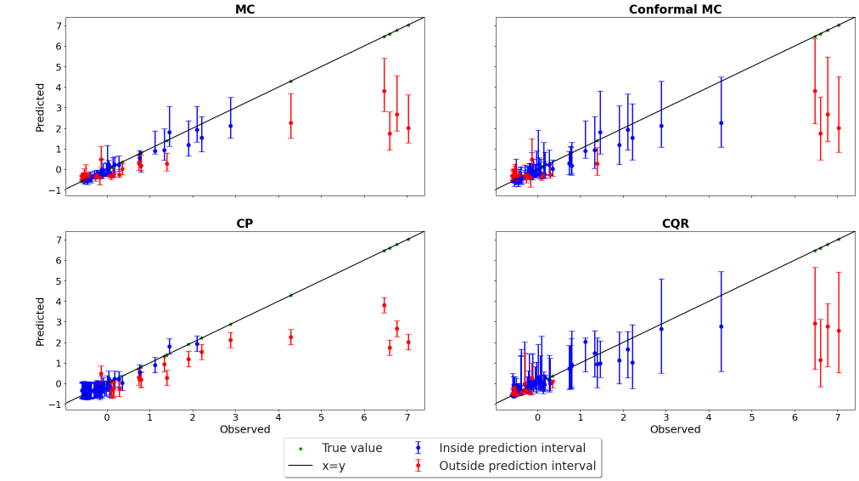}
\caption{Empirical 
$Y$-conditional coverage. MC prediction intervals severely undercover, standard CP produces constant-width PIs, while CQR and Conf-MC generate adaptive, variable-width intervals.}
\label{fig:conditional}
\end{figure*}

Table \ref{tab:res_2} reports the corresponding average interval width (AIW). While no single method consistently achieves both the desired coverage and minimal interval width across all datasets, ConfMC offers a balanced trade-off among different factors. It produces intervals that are not too wide while remaining adaptive (unlike standard CP), allowing it to better capture data heterogeneity and provide more informative uncertainty estimates. Moreover, it does not require retraining when adjusting the confidence level (unlike CQR). Figure \ref{fig:conditional} illustrates this behavior on the Kernel dataset: CP intervals are fixed in width, whereas ConfMC produces adaptive intervals that maintain correct coverage.
\begin{table*}[h!]
\caption{Average Empirical Coverage (AEC) for $\alpha = 0.2$ across the 20 runs.}
\centering
\setlength{\tabcolsep}{5pt} 
\renewcommand{\arraystretch}{1} 
\begin{tabular}{@{}lccccc@{}}
\toprule
                    & \textbf{CP} & \textbf{CQR} & \textbf{NGB} & \textbf{MC} & \textbf{ConfMC} \\ \midrule
\textbf{Concrete}   & 0.87  $\pm$ 0.03     & 0.81  $\pm$ 0.04  & 0.75  $\pm$ 0.04      & 0.67  $\pm$ 0.04      & 0.81  $\pm$ 0.04          \\
 \textbf{Kernel}     & 0.85   $\pm$ 0.03       & 0.80   $\pm$ 0.04      & 0.84   $\pm$ 0.03       & 0.53   $\pm$ 0.09     & 0.77   $\pm$ 0.05         \\
\textbf{Supercond.}    & 0.79  $\pm$ 0.05    & 0.81  $\pm$ 0.04   & 0.59 $\pm$ 0.04      & 0.45  $\pm$ 0.04      & 0.80  $\pm$ 0.04       \\
\textbf{Blog}       & 0.83 $\pm$ 0.08       & 0.81 $\pm$ 0.03        & 0.73 $\pm$ 0.04        & 0.50 $\pm$ 0.16       & 0.80 $\pm$ 0.03          \\
\textbf{Popularity} & 0.83  $\pm$ 0.05     & 0.74  $\pm$ 0.18      & 0.84  $\pm$ 0.03       & 0.68  $\pm$ 0.12      & 0.81 $\pm$ 0.04       \\
\bottomrule
\end{tabular}
\label{tab:res_1}
\end{table*}
\begin{table*}[h!]
\caption{Average Interval Width (AIW) for $\alpha = 0.2$ across the 20 runs.}
\centering
\setlength{\tabcolsep}{4.5pt} 
\renewcommand{\arraystretch}{1} 
\begin{tabular}{@{}lccccc@{}}
\toprule
                    & \textbf{CP} & \textbf{CQR} & \textbf{NGB} & \textbf{MC} & \textbf{ConfMC} \\ \midrule
\textbf{Concrete}   & 1.05  $\pm$ 0.11      & 0.76  $\pm$ 0.07     & 0.63  $\pm$ 0.04        & 0.61 $\pm$ 0.04       & 0.84  $\pm$ 0.09         \\
\textbf{Kernel}     & 0.65  $\pm$ 0.09       & 0.74 $\pm$ 0.10        & 0.85 $\pm$ 0.07        & 0.52  $\pm$ 0.10        & 0.93  $\pm$ 0.17            \\
\textbf{Supercond.}    & 1.12  $\pm$ 0.15     & 1.00 $\pm$ 0.09     & 0.54  $\pm$ 0.03       & 0.45 $\pm$ 0.04      & 1.09  $\pm$ 0.15          \\
\textbf{Blog}       & 0.32 $\pm$ 0.08      & 0.51 $\pm$ 0.13      & 0.19 $\pm$ 0.06        & 0.30 $\pm$0.09       & 0.64 $\pm$ 0.27           \\
\textbf{Popularity} & 0.91  $\pm$ 0.10       & 0.35 $\pm$ 0.08      & 0.02  $\pm$ 0.01         & 0.45  $\pm$ 0.05     & 0.62 $\pm$ 0.14          \\
\bottomrule
\end{tabular}
\label{tab:res_2}
\end{table*}

We conclude that ConfMC provides well-calibrated, adaptive prediction intervals. Furthermore, ConfMC does not require retraining when adjusting the confidence level, making it particularly suitable for rapid and flexible deployment. In summary, while all conformal-based methods achieve the desired coverage, ConfMC distinguishes itself by combining adaptive interval width with computational efficiency, providing both accurate and informative uncertainty estimates.

\begin{figure*}[h!]
\centering
\includegraphics[width=0.65\linewidth]{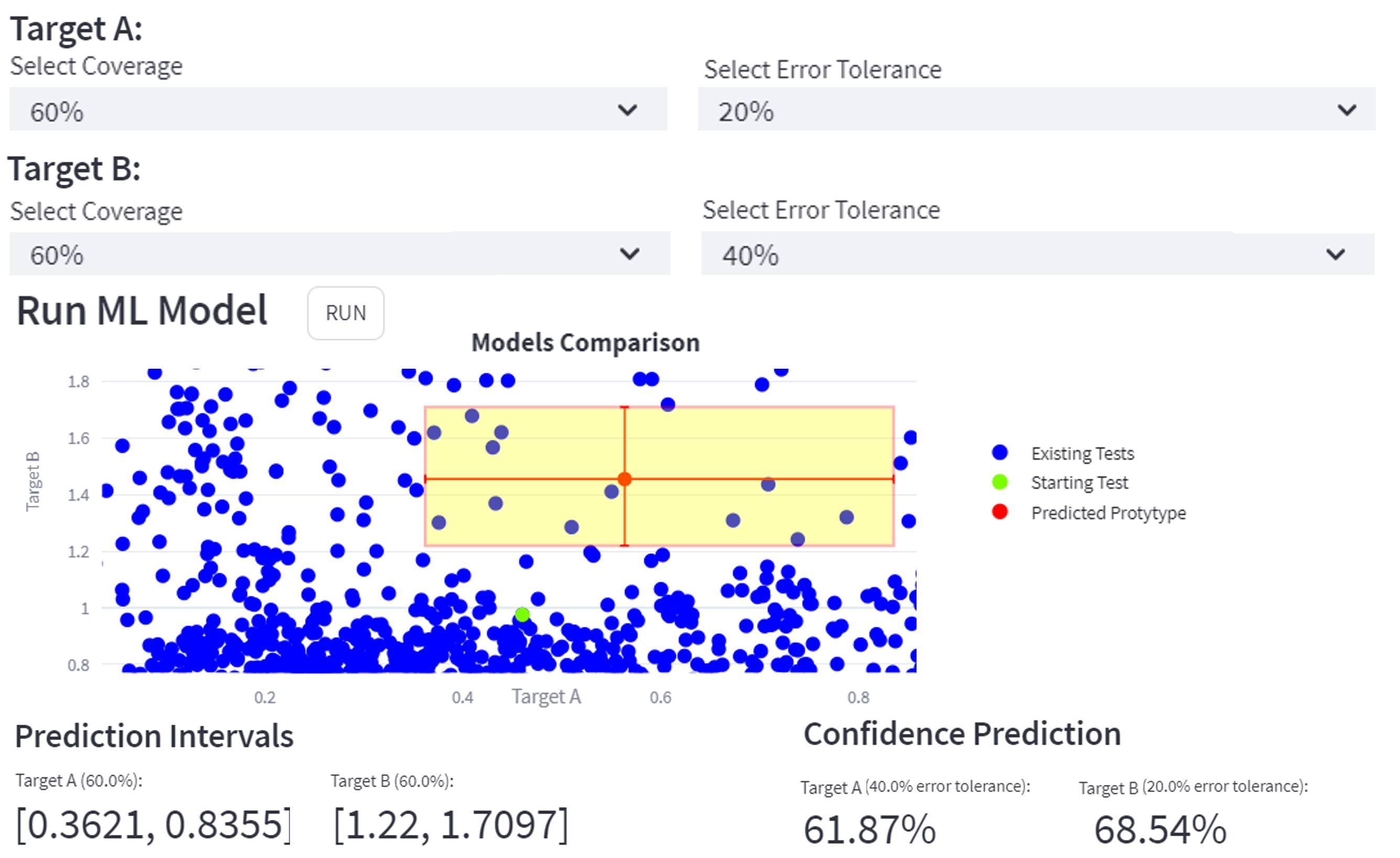}
\caption{Web interface implementing the proposed method with user-selected coverage and error tolerance. Blue dots: tested products; green dot: search start; red dot: search end; red rectangle: prediction intervals at the chosen error rate.}
\label{fig:stream}
\end{figure*}

\section{Industrial Application of Gradient based Search}
We implemented our solution within a leading construction tools manufacturing company to assess its practical effectiveness. We aimed to enhance product design workflows using machine learning techniques. The primary data source for this study was internal data collected from diamond segment prototypes. Each prototype was tested under a different set of environments. Overall, the data consists of design specifications of the prototype, testing conditions and the measured speed and lifetime of the prototype as the output variable.

We show a snapshot of how the Projected Gradient based search was deployed internally (Figure \ref{fig:stream}) using the \textit{Streamlit} platform. The algorithm was wrapped in a user-friendly web application that allows the design engineers to explore and create new product prototypes. Importantly, the prediction intervals show the user the uncertainty of the model's prediction about the new proposed prototype. 

User surveys demonstrated that the solution leads to fewer manufactured prototypes; resulting in faster product development and reduced wastage of resources. Moreover, uncertainty estimation greatly helped in gaining the trust of design engineers. This optimization method was one important element in a larger project that helped to drive data-driven product development. 

\section{Conclusions and Future Work}
We presented a robust framework for Data-Driven Product Development that uses neural networks to model the unknown mapping from product features to properties and leverages Projected Gradient Descent to identify optimal prototypes within user-defined feasible regions. By exploiting the differentiability of the network with respect to its inputs, Projected Gradient Descent efficiently searches the feasible space. Uncertainty quantification is provided via Monte Carlo Dropout in combination with Nested Conformal Prediction (ConfMC), delivering valid, adaptive, and user-controllable prediction intervals that support reliable and interpretable decision-making.

Our approach addresses multiple correlated targets and constrains extrapolation, but it assumes data exchangeability. In practice, feedback covariate shift \cite{cp_feedback}, signifying a shift in distribution during the search process, may occur. We attempt to mitigate this issue by including a projection step during the search, and by relying on the Monte Carlo predictive distribution, which is expected to produce wider intervals for points from distributions distinct from the training set. A thorough investigation of these assumptions and their implications remains an important avenue for future research. Incorporating automatically generated explanations could further enhance practical usability.

Overall, our method offers precise prototype optimization, actionable uncertainty estimates, and flexibility for user-guided exploration, while highlighting directions for extending its applicability in real-world design scenarios.

\section*{\uppercase{Acknowledgements}}
Generative AI tools (large language models) were used for editing purposes, such as grammar correction, spelling, and word choice refinement.

\bibliographystyle{apalike}
{\small
\bibliography{example}}

\end{document}